\documentclass[conference]{IEEEtran}
\usepackage{cite}
\usepackage{amsmath,amssymb,amsfonts}
\usepackage{algorithmic}
\usepackage{graphicx}
\usepackage{textcomp}
\usepackage{xcolor}
\usepackage{booktabs} % for nicer horizontal lines KK
\usepackage{siunitx} % for aligning numbers at the decimal point KK
\usepackage{url} % KK
\usepackage{makecell} % KK

\usepackage{array}    % for \rowstyle KK
\usepackage{colortbl} % optional, if you also want row coloring KK
\newcommand{\gattf}{GATTF} % KK

\def\BibTeX{{\rm B\kern-.05em{\sc i\kern-.025em b}\kern-.08em
    T\kern-.1667em\lower.7ex\hbox{E}\kern-.125emX}}
\begin{document}

\title{Geographically-aware Transformer-based Traffic Forecasting for Urban Motorway Digital Twins
\\
% {\footnotesize \textsuperscript{*}Note: Sub-titles are not captured in Xplore and should not be used}
}

%=====================================================KK
\author{\IEEEauthorblockN{Krešimir Kušić}
\IEEEauthorblockA{
% \textit{Trinity College Dublin}\\
% Dublin, Ireland \\
% \textit{Faculty of Transport \& Traffic Sciences} \\
\textit{University of Zagreb}\\
Zagreb, Croatia \\
kkusic@fpz.unizg.hr}
\and
\IEEEauthorblockN{Vinny Cahill}
\IEEEauthorblockA{
% \textit{School of Computer Science \& Statistics} \\
\textit{Trinity College Dublin}\\
Dublin, Ireland \\
vjcahill@tcd.ie}
\and
\IEEEauthorblockN{Ivana Dusparic}
\IEEEauthorblockA{
% \textit{School of Computer Science \& Statistics} \\
\textit{Trinity College Dublin}\\
Dublin, Ireland \\
ivana.dusparic@tcd.ie}
}
%===========================KK

% ================
\maketitle

\begin{abstract}
The operational effectiveness of digital-twin technology in motorway traffic management depends on the availability of a continuous flow of high-resolution real-time traffic data.
To function as a proactive decision-making support layer within traffic management, a digital twin must also incorporate predicted traffic conditions in addition to real-time observations. Due to the spatio-temporal complexity and the time-variant, non-linear nature of traffic dynamics, predicting motorway traffic remains a difficult problem. Sequence-based deep-learning models
offer clear advantages over classical machine learning and statistical models in capturing long-range, temporal dependencies in time-series traffic data, yet limitations in forecasting accuracy and model complexity point to the need for further improvements.
To improve motorway traffic forecasting, this paper introduces a Geographically-aware Transformer-based Traffic Forecasting (\gattf) model, which exploits the geographical relationships between distributed sensors using their mutual information (MI).
The model has been evaluated using real-time data from the Geneva motorway network in Switzerland and results confirm that incorporating geographical awareness through MI enhances the accuracy of \gattf~forecasting compared to a standard Transformer, without increasing model complexity.

% Vinny, Ivana
%%%%%%%%%%%%%%%%%%%%%%%%%%%%
% DT needs acurrate traffic forecast, fine granularity data ...
% Traffic forecast is hard...
% Sequence models can help...
% Long term dependences.. by seq. models.
% Improved by geo. aw.
% using Mutual information integrated with seq. model.
% Good results.
%%%%%%%%%%%%%%%%%%%%%%%%%%%%

%%%%%%%%%%%%%%%%%%%%%%%%%%%%%%%%%%%%%
% We propose a framework that integrates mutual information (MI) theory with transformer architectures for traffic forecasting, leveraging MI to identify a subset of the most informative covariates (traffic detectors), thereby improving the predictive accuracy of the target detector time series.

% inherent volatility

% to discern intricate dependencies and
% capture long-range relationships

% asymmetric traffic flow dynamics according to the specific road network topology and socio-economic traveller behaviors

% RNNs, LSTMs, faced challenges in capturing long-range dependencies and suffered from
% sequential processing inefficiencies.

% capture intricate patterns in sequential data

% Ablation Study

% ensuring models learn from structurally aligned clusters of sensors' time series

% LLM shortcomings
% Despite their remarkable performance, LLMs suffer from some significant drawbacks. First, they are trained on general-purpose data and have lower performance in domain-specific tasks and low- resource languages. 
% Secondly, they often reflect societal biases present in training data, which can result in biased outcomes. Third, LLMs sometimes produce inaccurate or made-up information, termed “hallucinations”.

\end{abstract}


\section{Introduction}

Constantly increasing mobility demand imposes a significant strain on the motorway systems with limited capacities.
% Existing traffic control methods have partially addressed this challenge.
Additionally, the emergence of connected and automated vehicles (CAVs) is gradually introducing vehicle heterogeneity. This requires new traffic control paradigms that meet future traffic needs and enable control over heterogeneous vehicles while supporting strategic CAV planning~\cite{2023Fan_CAV_DT}.
Therefore, traffic modeling and control require continuous advancement through new technological solutions that can efficiently leverage real-time traffic information from road units and vehicles.
These advanced methods utilize digital twins (DT) and artificial intelligence (AI) models to enable more accurate traffic analysis and control~\cite{2025Guo_LLM_driven_DT_for_ITS_Survey}. 

Moreover, effective traffic management on high-speed roads, such as urban motorways, requires not only an accurate estimate of current traffic conditions but also a prediction of future traffic~\cite{Zhang2024_Traffic_Forecasting_for_Freeway_DT}.  
Motorway DT leverages real-time data from diverse traffic sensors to provide a promising framework for real-time traffic monitoring and decision-making. By integrating real-time fine-grained traffic data with low-latency, DT bridges real-world processes into the digital \emph{virtual} space~\cite{2023_KUSIC_DT_GM}. By doing so, DT model allows for the safe optimization of control policies, thereby reducing the risk of implementing suboptimal or safety-critical control strategies in real traffic systems~\cite{2024_Irfan_DT_in_Transport_Review}.

To serve as a proactive decision-support layer in traffic management, DT must also incorporate means to accurately predict traffic states in addition to and based on current observations~\cite{2025Guo_LLM_driven_DT_for_ITS_Survey,2024_Irfan_DT_in_Transport_Review}.

Forecasting traffic states on urban motorways is difficult because of the nonlinear dynamics of traffic flow. Exogenous factors such as weather, incidents, and demand changes add to the challenge. Urban motorway traffic is strongly affected by inflows and outflows from local urban arterials. These local interactions introduce noise and uncertainty into demand, increasing the variability of motorway traffic state modalities~\cite {Li_2023_Mitigating_urban_mot_congest_via_active_TM}. As a result, there is no universal forecasting method. Forecast performance is highly location-specific, depending on demand patterns as well as the topology of the motorway and surrounding network~\cite{SHAO2025_CCDSReFormer}.
While traffic demand remains the primary factor for reliably inferring traffic states, accurately estimating and predicting it is challenging.

Sequence-based deep-learning models based on the Transformer architecture have demonstrated strong performance in time-series prediction~\cite{Qingsong_2023_Transformers_in_time_series_survey}. By learning autoregressive relationships and long-term dependencies across multiple steps (positions in the sequence) in time-series data, these models can capture complex temporal structure in data.
Such models are thus effective in capturing variations in
historical patterns of traffic data measured by traffic sensors~\cite{Zhang_2023_Pure_Transformer_for_Traff_Forecasting}.
However, sudden and rare spikes in traffic flow at specific motorway locations may not be well captured when the model is trained on data from a large set of spatially distributed sensors.
Incorporating all sensors can make the model too general, overlooking geographically specific traffic behavior and the daily fluctuations inherent to each site.
In contrast, training the model only on a specific site preserves the site-specific inherent characteristics but overlooks interactions with nearby traffic sensor locations and temporal patterns throughout the network, such as traffic lags, which are essential for robust traffic forecasting.
This creates a trade-off: using all sensors risks overgeneralization, while focusing on individual sites loses network-wide interactions.

To address these problems, this paper introduces a Geographically-aware Transformer-based Traffic Forecasting (\gattf) model, which exploits the geographical relationships between distributed sensors captured using their mutual information (MI)~\cite{Kraskov2004_estim_mutual_info}. MI detects non-linear correlations between sensors and captures variability and irregularities in traffic flow. By capturing shared information, MI also reflects road network topology features and temporal changes driven by flow dynamics. Using MI, \gattf~captures lagged influences from distributed sensors and finds the most informative ones. This preserves essential geographical context and avoids overgeneralization. MI is thus used to select sensors for constructing the covariates of the
traffic flow characterization, which are combined with other
inputs and processed by a transformer model. Consequently, modeling the non-linear covariate–target correlations identified by MI can substantially reduce uncertainty in motorway traffic forecasting by enriching the feature space with the most informative signals while maintaining constant model capacity.

In prior work, MI has primarily been used as a metric for evaluating input–output information gain in sequence modeling, while its potential to systematically uncover spatio-temporal dependencies among motorway sensor measurements remains largely unexplored, particularly for identifying the most informative, geographically aware covariates for Transformer-based forecasting models.

This study makes the following contributions:
\begin{itemize}
   \item It proposes a geographically-aware transformer-based model for traffic forecasting on an urban motorway.
   \item It applies MI as a feature selector to capture nonlinear spatio-temporal dependencies among sensors, enriching the Transformer’s features with informative, geographically relevant covariates for better traffic forecasting.
\end{itemize}

The \gattf~model has been evaluated on real-time traffic data from the Geneva motorway network in Switzerland, demonstrating improved traffic forecasting performance compared to baseline scenarios.

The remainder of this paper is organized as follows.
Section II discusses related work. Section III provides the necessary background on MI. Section IV describes our transformer architecture for traffic forecasting. Section V explains how we enhance it for geographically-aware traffic forecasting. Section VI presents results and analysis. Conclusions are given in Section VII.

\section{Related work}

While there is a long history of research into traffic flow prediction~\cite{SHAYGAN2022_TrafficPrediction_using_AI_review}, recent research has particularly investigated the application of sequence-based deep learning models to forecast the evolution of traffic flows.
Among these, the Transformer, originally designed for sequence-to-sequence machine translation~\cite{2017_Vaswani_AttentionIsAllYouNeed}, has been adapted for time-series modeling and used for various forecasting tasks. 

Its self-attention mechanism captures long-range temporal dependencies. It also accounts for traffic's inherent nonlinearity (as seen in the data) through a deep learning architecture~\cite{2020Cai_Traffic_Transformer}.

While effective at modeling long-range temporal dependencies, these models inadequately capture spatial dependencies in traffic data, revealing the need for hybrid approaches~\cite{Qingsong_2023_Transformers_in_time_series_survey}.

For example, in~\cite{2020Cai_Traffic_Transformer}, the Traffic Transformer model is proposed. By extending Google’s Transformer framework with a graph convolutional neural network (GCNN), model is able to partially account for spatial dependencies among time series. However, their approach relies solely on a temporal attention mechanism. The influence of upstream road regimes on downstream traffic may vary over time in a nonlinear fashion, which underscores the need for a model with an architecture capable of capturing spatio-temporal traffic context, e.g., a spatio-temporal attention mechanism.

To capture spatio-temporal features of traffic,~\cite{2025Chang_Trafficformer_short_term_forecasting_considering_traffic_spatiotemporal_correlation} proposed the Trafficformer model for traffic speed prediction. A matrix encoding spatio-temporal traffic flow characteristics, by applying a spatial mask that defines strongly and weakly correlated nodes based on travel time, was provided as input to the transformer. The proposed model outperformed both classical linear methods and commonly used non-linear approaches.~\cite{TraffPrediction_MI_Clustering_2023_Huang} proposed TrafficTL, a cross-city transfer learning framework designed to improve traffic forecasting in data-scarce cities by leveraging data from data-rich ones. Its core innovation is the Temporal Cluster Block, which groups traffic sensors across source and target cities by maximizing MI between different temporal views of their time series, ensuring that clustered nodes share highly informative and synchronized traffic patterns. 
This MI-driven clustering reduces negative transfer by aligning only relevant data sources with the prediction model, thereby significantly improving forecasting performance.
In~\cite{yamaguchi2025citras_covariate_informed_transformer_time_series}, a patch-based, decoder-only Transformer is introduced. This model demonstrated the usefulness of integrating informative past and future covariates as inputs while maintaining autoregressive modeling. It demonstrated its effectiveness across thirteen real-world datasets, including traffic forecasting.
Similarly, the Temporal Fusion Transformer~\cite{lim2020temporal_fusion_transformers_known_unknown_covariates} leverages additional information. It incorporates static covariate inputs, which encode time-invariant known features, and observed historical covariates. These enrich the feature space and improve learning.

In summary, existing studies differ mainly in how transformers are augmented. Some supplement the architecture by integrating models such as GCNNs. Others provide additional covariate inputs to capture spatial traffic relationships. However, the systematic identification and integration of informative covariates to enrich the geographical awareness of transformer-based traffic forecasting models for urban motorways has rarely been addressed.

Motivated by the detected gap, this work investigates how MI can be leveraged to systematically identify strong correlations between distributed traffic sensors. This approach augments the transformer's input space with additional informative covariates. These covariates capture non-linear, lagged relationships among spatially distributed sensors, helping identify a subset that can improve the prediction of weakly predictable traffic sensor data (targets) across an urban motorway.

Such enriched feature representations give the Transformer geographical awareness, i.e., additional contextual traffic information. This enables the model to better capture spatio-temporal dependencies in traffic data caused by complex intrinsic traffic dynamics and network topology specific to urban motorways. Thus, the proposed \gattf~framework is particularly relevant for traffic flow prediction, where distributed sensor observations reflect spatial heterogeneity of traffic regimes governed by independent local demand patterns and network topology.

\section{Mutual Information}
 
MI is a concept from information theory that quantifies the amount of information that one random variable contains about another~\cite{Kraskov2004_estim_mutual_info}. Formally, for two discrete random variables $X$ and $Y$ with the joint distribution $p(x,y)$ and the marginal distributions $p(x)$ and $p(y)$, their mutual information, $I(X,Y)$, is defined as:

\begin{equation}\label{Eq:mutual_information}
I(X,Y)
\;=\;
\sum_{x\in\mathcal{X}} \sum_{y\in\mathcal{Y}} p(x,y)\,
\log\!\left(\frac{p(x,y)}{p(x)\,p(y)}\right)
\end{equation}
% \noindent where...(short)

To estimate MI among traffic time series, the data must be discretized into bins to compute marginal and joint probabilities, which are then used to calculate MI.
To define the proper number of bins (bin width), the Freedman–Diaconis rule has been used~\cite{freedman1981histogram}:
\begin{equation}
h = 2 \frac{\mathrm{IQR}(x)}{n^{1/3}},
\end{equation}

\noindent where $h$ is the bin width, $\mathrm{IQR}(x)$ is the interquartile range of the data and $n$ is the number of observations. 
The number of bins $k$ can then be computed as:

\begin{equation}
k = \left\lceil \frac{\max(x) - \min(x)}{h} \right\rceil.
\end{equation}

In the context of traffic time series, MI quantifies how much information the measurements of one sensor provide about the measurements of another sensor. For example, considering sensors at locations A6 and B2 in Fig.~\ref{fig:motorway_model_informative_detectors}, MI measures how much knowing the traffic at B2 reduces uncertainty about traffic at A6, capturing the variability in traffic flow. Sensors with higher MI regarding target sensors can therefore be considered more informative covariates to be used as additional inputs in a transformer model for improving prediction accuracy.

\section{\gattf~Transformer Architecture for Traffic Forecasting}

As discussed in Section II, Transformers have emerged as a promising approach for capturing long-range dependencies and complex temporal patterns in traffic~\cite{Qingsong_2023_Transformers_in_time_series_survey}. In this section we outline salient details of the Transformer architecture used in this work.

\subsection{Transformer architecture}
The original Transformer~\cite{2017_Vaswani_AttentionIsAllYouNeed}, developed for natural language processing, uses an encoder–decoder architecture in which both components consist of stacked blocks. Each encoder block contains a multi-head self-attention layer followed by a feed-forward network, while each decoder block also includes a cross-attention layer that connects encoder and decoder representations. This encoder–decoder design is especially useful at inference time, when logged data are used to forecast a sequence of future steps (Fig.~\ref{fig:MI_feature_selector}a).

When adapting Transformers for time series modeling, several architectural modifications have been introduced at both the module and system levels to enhance their suitability. Instead of fixed sinusoidal positional encodings, we use a model with timestamp-based embeddings that capture seasonality and periodic cycles (e.g., daily, weekly, yearly)~\cite{Qingsong_2023_Transformers_in_time_series_survey}. Moreover, since time series can be very long, it is computationally infeasible to input the full history at once due to the quadratic cost of attention. To address this, we apply a model that is trained on context windows of appropriate length, paired with a prediction window. The encoder processes the context window, while a causal-masked decoder predicts the future sequence step by step, similar to teacher forcing in sequence-to-sequence models. Finally, we use a model whose inputs are augmented with lagged values sampled from different past time horizons, which serve as contextual signals, much like keywords in text, helping the model learn temporal dependencies more effectively. This enables the Transformer to effectively learn from long sequences while respecting temporal order~\cite{2020_wolf_huggingfacestransformersstateoftheartnatural}.

Beyond point forecasting, many real-world applications, including traffic management, require models that quantify prediction uncertainty. Probabilistic forecasting addresses this need by learning a distribution over future outcomes rather than a single deterministic value. 
To address this, we use a Transformer-based probabilistic time series model, implemented in~\cite {2020_wolf_huggingfacestransformersstateoftheartnatural}, that leverages deep learning’s capacity to learn shared representations across multiple related time series while modeling uncertainty. This capability is particularly advantageous for DT-assisted motorway traffic management, where uncertainty-aware forecasts support the design of control strategies that remain robust under a variety of future traffic conditions.

\subsection{Transformer implementation}
For \gattf~time series forecasting, we leverage an open-source Transformer implementation from Hugging Face, introduced in the blog post \emph{Probabilistic Time Series Forecasting with Transformers}\footnote{https://huggingface.co/blog/time-series-transformers}~\cite{2020_wolf_huggingfacestransformersstateoftheartnatural}. The model parameters used in the work are defined as follows.
The prediction length is 288 (corresponding to one day of 5-minute intervals), and the context length is 576. The lag sequence length and time features are determined by the data sampling frequency (5 minutes).
The transformer architecture consists of four encoder layers and two decoder layers, with a model dimension of 256. Both the encoder and decoder use eight attention heads, and the feed-forward network dimensions in the encoder and decoder are set to 1024.

\subsection{Dataset description}

In this work, fine-grained, real-time traffic data from counters (sensors) installed along major sections of the Geneva motorway were used~\cite{2023_KUSIC_DT_GM}. Counters record traffic by direction and time, are updated every minute, and are accessible via the \emph{opentransportdata.swiss} API\footnote{https://opentransportdata.swiss/en/rt-road-traffic-counters/}. We aggregated the data to a 5-minute resolution, resulting in 19 days of lane-level traffic flow measurements in vehicles per hour (veh/h) for passenger cars. Each time series, thus, corresponds to one sensor (containing 5472 measurements), and the dataset comprises 14 series from counters at locations A, B, and C in Fig.~\ref{fig:motorway_model_informative_detectors}. Additionally, the examined period does not include holidays, and information on spatial events or major incidents is not currently provided via the API in real time. The dataset contains no missing values. However, if missing values occur, the transformer model masks them so they do not contribute to loss computation or model parameter updates during training. The model does not normalize data during preprocessing, instead, it internally normalizes embeddings and hidden features, which helps stabilize learning and predictions. The lengths of the time series are 4,896, 5,184, and 5,472 for training, validation, and testing, respectively.

\section{Geographically-aware Traffic Forecasting}

Building on the complementary strengths of probabilistic time series modeling and MI, in this section, we introduce \gattf, which leverages the transformer’s flexible input representation and ability to capture long-term temporal dependencies, along with MI’s capacity to quantify spatio-temporal relationships among sensors. This enables the \gattf~to identify the most informative covariates and augment its inputs for improved forecasting accuracy.

\subsection{Spatio-temporal dependencies in traffic flow}
Traffic sensors often exhibit strong spatial coherence as the unidirectional flow of vehicles along the motorway creates dynamic dependencies between upstream and downstream locations, while disturbances such as shock waves can propagate in both directions. Although upstream flow can inform predictions of downstream traffic (considering directional lag effects), relying solely on immediate upstream neighbors is often insufficient. For example, as shown in Fig.~\ref{fig:motorway_model_informative_detectors}, sensors B1, B2, B3, B4, A4, A5, and A6 all provide valuable information for predicting C3 (confirmed by mutual information values shown in Table.~\ref{tab:mutual_information}), illustrating that upstream-downstream dependencies are neither strictly local nor directly aligned, since B1 and B2, for instance, are not directly upstream of C3. 
Their signals influence C3 indirectly via traffic merging at the major grade-separated interchange, particularly by the number of vehicles heading to the C34–A456 branch from the opposite motorway direction. 
Thus, identifying the most informative sensor locations is not straightforward, even in this relatively simple motorway network. Not all sensors contribute equally as sources of useful information for a given subset of target sensors. 
Therefore, systematically identifying the subset of sensors that provide meaningful information about the target group requires a methodical approach.

Furthermore, when learning dependencies in sequence data with transformers, traffic measurements recorded by sensors are influenced not only by their site-specific traffic conditions but also by neighboring traffic conditions captured by neighboring sensors. Transformers are well-suited to handle multiple input sequences, and by incorporating informative covariates alongside the target sensor’s data, the attention mechanism can learn which features are most relevant at each timestep. In this way, covariates serve as guiding signals that help the model distinguish recurrent traffic patterns from irregular events, capture inter-sensor interactions, and improve the predictive power of target sensor forecasts.

This underscores the value of using MI that is able to systematically capture nonlinear dependencies across the network sensors and select the most informative sensors to be used as covariates to enhance the Transformer input and thus improving the forecasting performance of target sensors, as detailed in the following sections.

\subsection{Geographically-Aware Covariate Selection}

Although the urban motorway model in Fig.~\ref{fig:motorway_model_informative_detectors} may appear simple at first glance, its traffic dynamics are inherently complex. The network includes a major grade-separated interchange with junctions to the east (city center), south (border with France), and north (toward Geneva Airport), creating highly non-linear traffic interactions. Additionally, multiple on- and off-ramps further increase forecasting uncertainty along the mainline. The system is also shaped by asymmetric demand patterns driven by daily commuter flows between France and Geneva, which amplify spatio-temporal variability in traffic conditions.

To enhance the transformer model with geographical awareness we include time-series data from the most informative spatially related sensors as inputs into the model as motivated above. To identify the most informative sensors \gattf~ incorporates an MI-based feature selection module (Fig.~\ref{fig:MI_feature_selector}c) that quantifies covariate–target dependencies across connected traffic sensors in the urban motorway network. 
Covariates provide the transformer model with contextual, temporal, and external information that enables it to model complex time series patterns more accurately. Without covariates, the model relies only on past values of the target variable, which can limit performance, particularly when changes in the series are driven by external conditions~\cite{lim2020temporal_fusion_transformers_known_unknown_covariates}.

\subsubsection{Integrating Spatial Awareness through MI-Based Covariates}

\begin{figure}[t]
     \begin{center}
        \includegraphics[width=0.6\columnwidth]{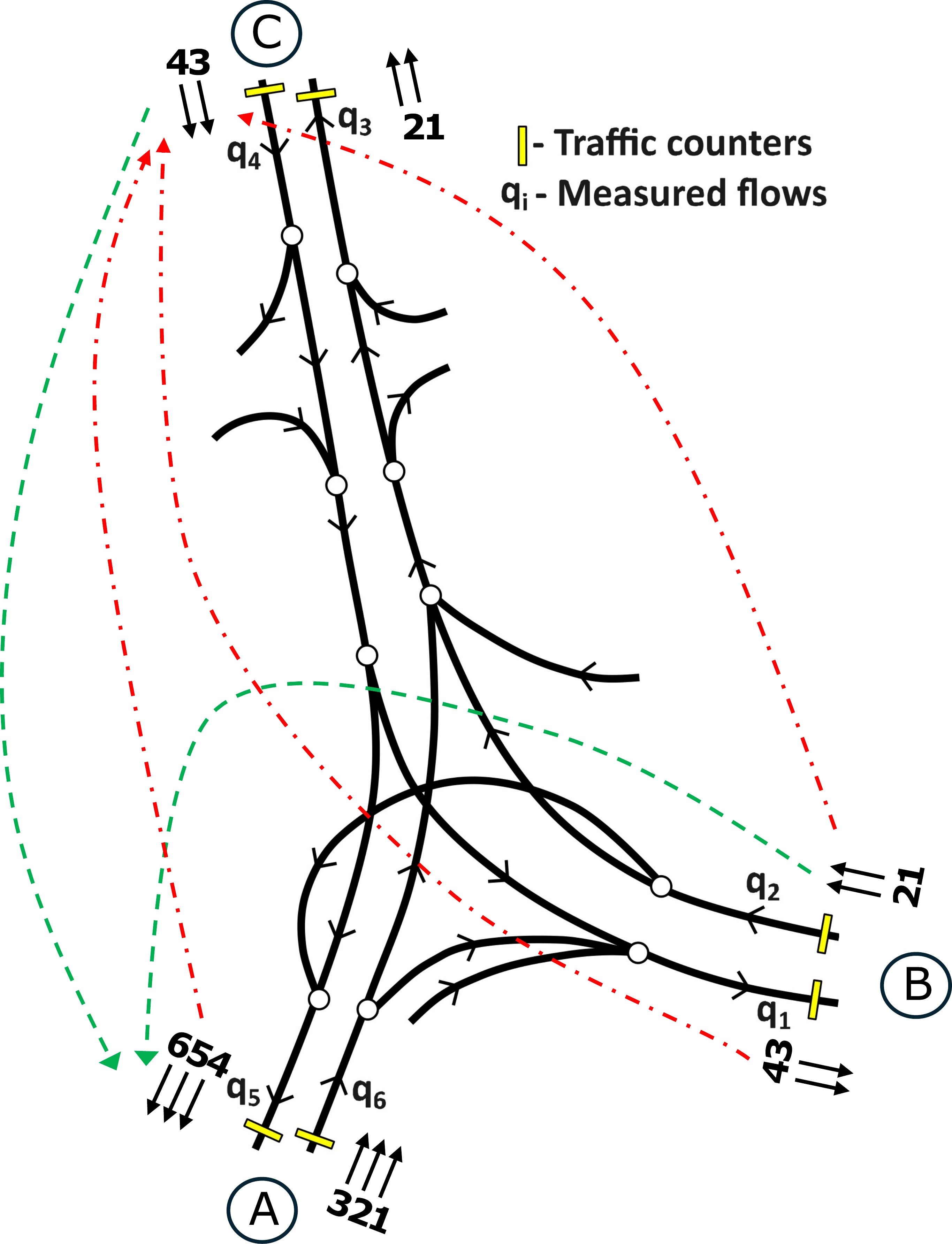}
          \caption{Graph representation of the Geneva motorway showing the most informative detectors for reducing forecasting error of e.g., A6, C3 detectors}
        \label{fig:motorway_model_informative_detectors}
        \end{center}
\end{figure}

\begin{figure}[t]
     \begin{center}
        \includegraphics[width=0.9\columnwidth]{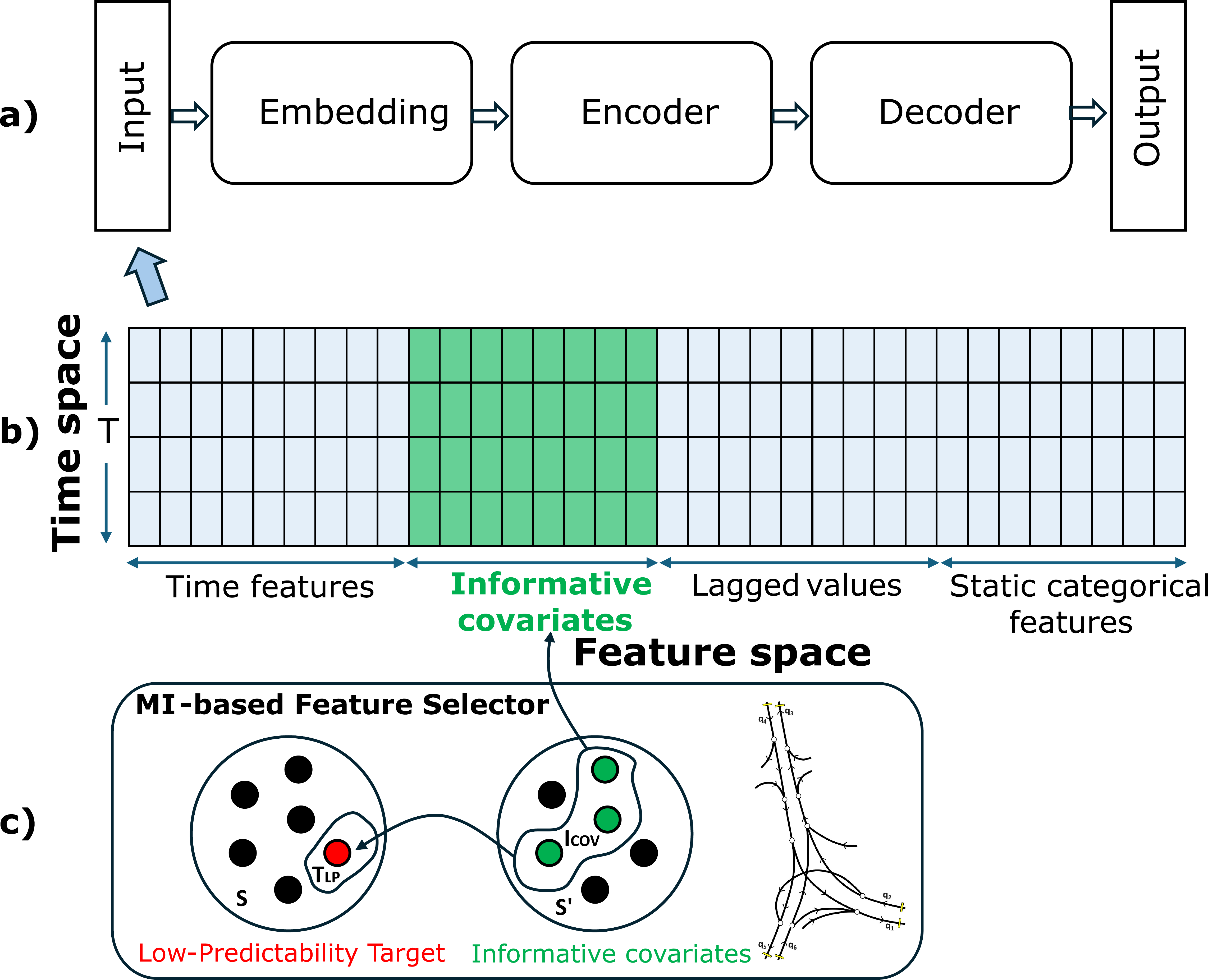}
          \caption{Geographically-aware input augmentation through MI-based informative covariate selection}
        \label{fig:MI_feature_selector}
        \end{center}
\end{figure}

The MI module selects a subset of sensors from which it constructs informative covariates of the spatially related traffic flow. This module does so by identifying covariates that convey the highest mutual information with low-predictability sensors (ones that are poorly predicted while the transformer is trained on the full set of sensors). 
Building on this general idea, we now define the problem more concretely.
When trained on the full set of network sensors $S$ (see Fig.~\ref{fig:MI_feature_selector}c), the transformer model may overgeneralize and fail to capture site-specific dynamics, making the subset of sensors $T_{LP}$ difficult to predict. To address this, the MI module uses mutual information to identify a subset of informative covariates $I_{COV}$ that improves the model’s ability to predict these low-predictability sensors $T_{LP}$, where $T_{LP} \subset S$ and $I_{COV} \subset S$ such that $T_{LP} \cap I_{COV} = \varnothing$.
This allows the encoder inputs to be conditioned with geographically-relevant signals (Fig.~\ref{fig:MI_feature_selector}b and~\ref{fig:MI_feature_selector}c). This targeted-covariate embedding improves the representation of spatio-temporal dependencies among sensors, thereby enhancing the Transformer’s forecasting performance at motorway locations with inherently complex dynamics. 
For illustrative clarity, in Fig.~\ref{fig:MI_feature_selector}c, we denote two sets $S$ and $S'$, although they represent the same set. The full set of sensors is $S = \{A_1, \dots, A_6, B_1, \dots, B_4, C_1, \dots, C_4\}$ (see Fig.~\ref{fig:motorway_model_informative_detectors}).

While a transformer model for time series is typically designed to be trained on all sensor data simultaneously, capturing coupled effects of traffic flow across the analyzed network and predicting each output, this approach often leads to over-generalization, which degrades predictions at locations with distinct traffic dynamics. For example, removing sensors with low MI values from S and including covariates with high MI improves prediction accuracy for sensors at locations A6 and C3, which were initially affected by over-generalization when the transformer was trained on the full set S. In contrast, training a model on a single sensor ignores the broader spatio-temporal context of motorway traffic flow. The MI module in \gattf~addresses this trade-off by systematically identifying the most informative covariates from neighboring sensors. For instance, for sensor A6, the most informative covariates (with the highest MIs) are shown in Table~\ref{tab:mutual_information} and depicted by dashed arrows in Fig.~\ref{fig:motorway_model_informative_detectors}. By conditioning the transformer on this geographically relevant subset, the model retains essential spatial context while mitigating over-generalization and improving forecasts at low-predictability locations.

\subsubsection{MI-Based Selection and Covariate Encoding}
In our approach, the MI-based selection of the most informative covariates is performed as a preprocessing step when constructing the model input dataset. Once the covariate set is fixed, the remainder of the training pipeline (i.e., instance splitting, data transformations, batch generation, and model fitting) is handled automatically by the transformer. A technical detail worth noting is that covariates are encoded differently from the standard Transformer setup, as only past-known covariates are available (future traffic conditions are unknown). Specifically, the \emph{past time features} and \emph{future time features} tensors are masked along the time axis using binary indicators (1s and 0s) to inform the model as to which covariate values are known and which correspond to future, unseen time steps.

In summary, in addition to temporal features such as minute of hour, hour of day, day of week, age, etc., as well as categorical features like sensor ID and lagged values (which provide temporal context), \gattf~incorporates MI-selected informative covariates as auxiliary inputs to capture spatial traffic dependencies, effectively conditioning the model with geographical awareness.

\section{RESULTS AND ANALYSIS}
To assess the impact of augmenting a Transformer model with MI-based informative covariates, we conducted an ablation study comparing: (i) \gattf~with the most informative covariates, (ii) \gattf~with less informative covariates (defined by mutual information $I$ in Table~\ref{tab:mutual_information}), (iii) a Transformer without covariates trained on the full set of sensors, and (iv) a Transformer without covariates trained on data from a single sensor. Performance metrics for time-series forecasting are categorized into two groups. The first group is scale-independent and includes the Mean Absolute Scaled Error (MASE) and the symmetric Mean Absolute Percentage Error (sMAPE).
To provide a more interpretable measure in the original data units, we also evaluate prediction accuracy using the Mean Absolute Error (MAE). Because most errors occur during traffic peaks, we also use the Root Mean Squared Error (RMSE), which penalizes larger deviations more than MAE, to assess the effectiveness of the prediction methods~\cite{2025Chang_Trafficformer_short_term_forecasting_considering_traffic_spatiotemporal_correlation}.

\subsubsection{Descriptive Analysis}
\begin{figure}[t]
     \begin{center}
        \includegraphics[width=1\columnwidth]{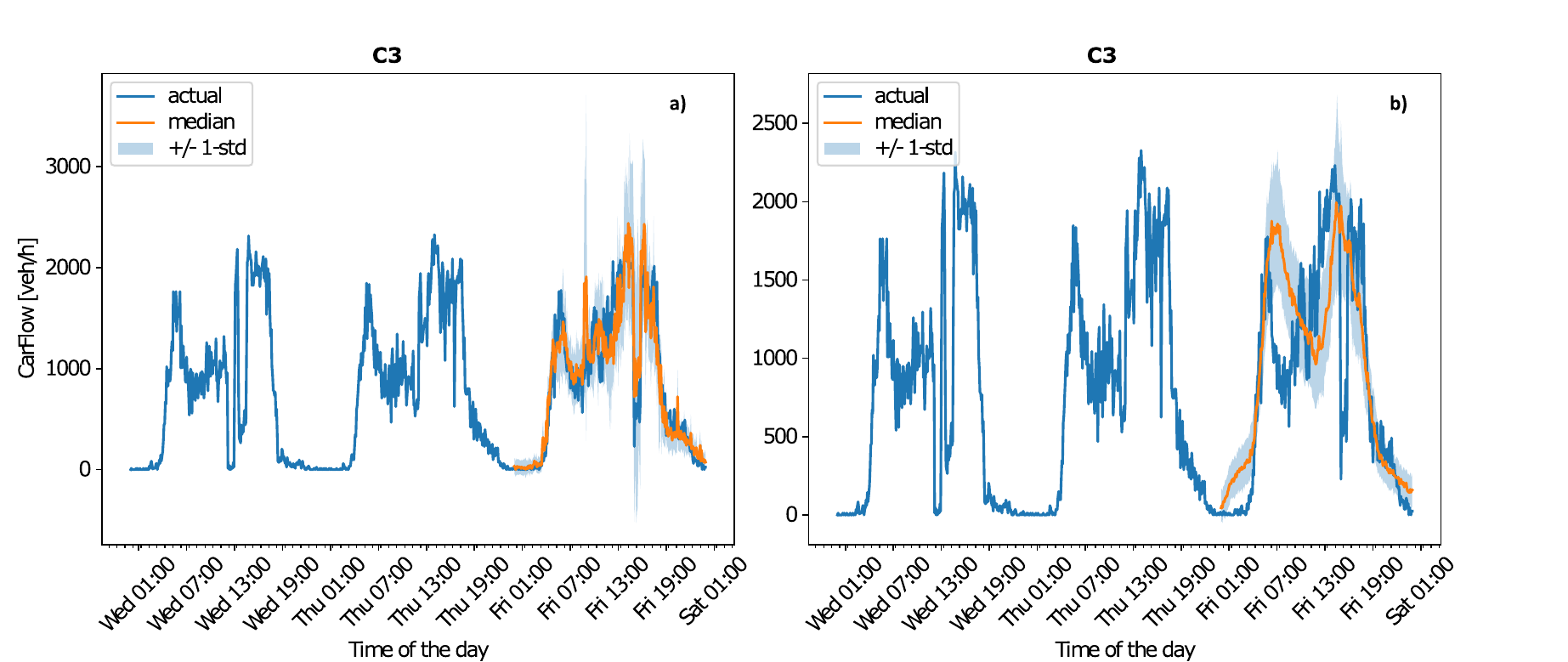}
          \caption{24-hour, 5-minute resolution traffic forecast at location C3: (a) \gattf~augmented with informative covariates, (b) standard Transformer using all sensors without covariates.}
        \label{fig:C3_prediction}
        \end{center}
\end{figure}

In Fig.~\ref{fig:C3_prediction}a and~\ref{fig:C3_prediction}b, the actual traffic (ground truth) is shown in blue, while the predicted traffic from the analyzed Transformer architectures is shown in orange. The forecasts correspond to Friday, May 20, 2022, at lane location C3. 
For \gattf~(Fig.~\ref{fig:C3_prediction}a), the predicted curve closely follows the ground truth, with only minor deviations at sharp peaks, demonstrating satisfactory predictive performance. The model accurately captures the morning and afternoon rush-hour peaks, and the sudden, sharp drop in traffic volume during the afternoon rush hour, as well as the traffic volume before and after the peaks.
In contrast, the standard Transformer (Fig.~\ref{fig:C3_prediction}b) shows noticeable discrepancies from the actual traffic throughout the prediction horizon. Detailed performance metrics are provided in Tables~\ref{tab:forecast_A6} and~\ref{tab:forecast_C3}.

\subsubsection{Comparison of traffic prediction performance}

Tables~\ref{tab:forecast_A6} and~\ref{tab:forecast_C3} show the prediction performance of the different approaches used in our study in terms of 24-hour-ahead prediction at locations A6 and C3, respectively.

Several noteworthy observations emerge from the results. Overall, the \gattf~model, whether using the most informative or less informative covariates consistently outperforms baselines that do not incorporate covariates. This demonstrates the effectiveness of MI in selecting covariates that capture relevant information about the target, efficiently enriching the model’s feature space and enabling it to account for the joint spatio-temporal effects of surrounding traffic when predicting future target values, without increasing model size.

For example, when predicting traffic at location C3, \gattf~with informative covariates achieves an 85.63\% improvement in MASE compared to a Transformer trained on all sensors without covariates.

A second interesting observation is that whether the model is trained on the full set of sensors to predict, for example, A6, or trained solely on A6, the prediction metrics are very similar.

\subsubsection{Mutual information analysis}

The MI analysis confirms strong spatial coherence among traffic sensors, reflecting complex upstream–downstream dependencies along the motorway. As shown in Table~\ref{tab:mutual_information}, sensors B1, B2, B3, B4, A4, A5, and A6 exhibit notably high MI values with sensor C3, indicating that C3's traffic dynamics are influenced by multiple, not strictly adjacent, locations. This demonstrates that traffic interactions extend beyond immediate upstream neighbors, with indirect effects such as merging flows at the major grade-separated interchange significantly shaping traffic dynamics on the analyzed urban motorway network. 

\begin{table}[h!]
\centering
\caption{Pairwise mutual information between sensors}
\label{tab:mutual_information}
% \begin{tabular}{lllS[table-format=2.4]S[table-format=2.4]S[table-format=1.4]S[table-format=1.4]}
\begin{tabular}{lcccccc}
\toprule
% Joint entropy
% Mutual informatio
% Entropy of target
% Entroipes of Ys
{} & {$X_{A6}$} & {$Y_{item}$} & {$I(X_{A6},Y)$} & {$X_{C3}$} & {$Y_{item}$} & $I(X_{C3},Y)$ \\
\midrule
% \rowstyle{\bfseries}
{1} & \textbf{A6} & \textbf{B1} & \textbf{1.5237} & \textbf{C3} & \textbf{B1} & \textbf{1.0747} \\
{2} & \textbf{A6} & \textbf{B2} & \textbf{1.3214} & \textbf{C3} & \textbf{B2} & \textbf{1.1781} \\
{3} & A6 & B3 & 0.9101 & \textbf{C3} & \textbf{B3} & \textbf{0.9644} \\
{4} & A6 & B4 & 0.9349 & \textbf{C3} & \textbf{B4} & \textbf{1.0605} \\
{5} & A6 & A1 & 0.7067 & C3 & A1 & 0.5262 \\
{6} & A6 & A2 & 0.7898 & C3 & A2 & 0.7071 \\
{7} & A6 & A3 & 0.5563 & C3 & A3 & 0.1747 \\
{8} & \textbf{A6} & \textbf{A4} & \textbf{1.4278} & C3 & A4 & 0.7649 \\
{9} & \textbf{A6} & \textbf{A5} & \textbf{1.3563} & \textbf{C3} & \textbf{A5} & \textbf{0.9099} \\
{10} & A6 & C1 & 0.9052 & C3 & A6 & 0.5695 \\
{11} & A6 & C2 & 1.0606 & C3 & C1 & 0.8746 \\
{12} & \textbf{A6} & \textbf{C3} & \textbf{1.6071} & C3 & C2 & 0.8832 \\
{13} & \textbf{A6} & \textbf{C4} & \textbf{1.5833} & \textbf{C3} & \textbf{C4} & \textbf{1.5103} \\
%%%%%%%%%%%%%%%%%%%%%%%%%%%%%%%%%%%%%%%%%%%%%%
% \textbf{{1}} & \textbf{CH:0200.06} & \textbf{CH:0224.01} & \textbf{4.3646} & \textbf{5.7750} & \textbf{5.775} & \textbf{1.5237} \\
% \textbf{{2}} & \textbf{CH:0200.06} & \textbf{CH:0224.02} & \textbf{4.3646} & \textbf{5.3963} & \textbf{5.396} & \textbf{1.3214} \\
% {3} & CH:0200.06 & CH:0224.03 & 4.3646 & 4.9558 & 4.956 & 0.9101 \\
% {4} & CH:0200.06 & CH:0224.04 & 4.3646 & 5.7175 & 5.718 & 0.9349 \\
% {5} & CH:0200.06 & CH:0200.01 & 4.3646 & 5.3794 & 5.379 & 0.7067 \\
% {6} & CH:0200.06 & CH:0200.02 & 4.3646 & 5.5622 & 5.562 & 0.7898 \\
% {7} & CH:0200.06 & CH:0200.03 & 4.3646 & 3.8012 & 3.801 & 0.5563 \\
% \textbf{{8}} & \textbf{CH:0200.06} & \textbf{CH:0200.04} & \textbf{4.3646} & \textbf{5.1779} & \textbf{5.178} & \textbf{1.4278} \\
% \textbf{{9}} & \textbf{CH:0200.06} & \textbf{CH:0200.05} & \textbf{4.3646} & \textbf{5.8303} & \textbf{5.830} & \textbf{1.3563} \\
% {10} & CH:0200.06 & CH:0272.01 & 4.3646 & 5.8858 & 5.886 & 0.9052 \\
% {11} & CH:0200.06 & CH:0272.02 & 4.3646 & 5.4556 & 5.456 & 1.0606 \\
% \textbf{{12}} & \textbf{CH:0200.06} & \textbf{CH:0272.03} & \textbf{4.3646} & \textbf{5.4105} & \textbf{5.411} & \textbf{1.6071} \\
% \textbf{{13}} & \textbf{CH:0200.06} & \textbf{CH:0272.04} & \textbf{4.3646} & \textbf{5.9197} & \textbf{5.920} & \textbf{1.5833} \\
\bottomrule
\end{tabular}
\end{table}

\begin{table}[h!]
\centering
% (5-min. resolution)
\caption{Performance of 24-hour forecasts for A6 under different input settings: baseline (all detectors, no covariates), A6-only, and A6-covariate-augmented}
\label{tab:forecast_A6}
\begin{tabular}{lcccc}
\toprule
Metric & \makecell{Transformer\\ all-sensors\\ (no cov.)} & \makecell{Transformer\\ A6-sensor\\ (no cov.)} & \makecell{Informative\\ cov. (\gattf)} & \makecell{Less informative\\ cov. (\gattf)}\\
\midrule
MASE        & 1.893   & 1.823   & \textbf{0.867}    & 1.413\\
sMAPE       & 0.867   & 0.793   & \textbf{0.663}    & 0.69\\
MAE         & 175.987 & 169.46  & \textbf{80.356}   & 131.193\\
RMSE        & 276.18  & 289.803 & \textbf{126.593}  & 227.463\\
\bottomrule
\end{tabular}
\end{table}

\begin{table}[h!]
\centering
\caption{Performance of 24-hour forecasts for C3 under different input settings: baseline (all detectors, no covariates), C3-only, and C3-covariate-augmented}
\label{tab:forecast_C3}
\begin{tabular}{lcccc}
\toprule
Metric & \makecell{Transformer\\ all-sensors\\ (no cov.)} & \makecell{Transformer\\ C3-sensor\\ (no cov.)} & \makecell{Informative\\ cov. (\gattf)} & \makecell{Less informative\\ cov. (\gattf)}\\
\midrule
MASE        & 1.485 & 1.245 & \textbf{0.8} & 0.955\\
sMAPE       & 0.74 & 0.615 & \textbf{0.43} & 0.505\\
MAE         & 338.035 & 282.475 & \textbf{182.26} & 216.685\\
RMSE        & 450.925 & 406.1 & \textbf{256.07} & 343.52\\
\bottomrule
\end{tabular}
\end{table}

\section{Conclusions and future work}

This paper introduces the Geographically-Aware Traffic Flow Forecasting model \gattf, which leverages the sequence-based modeling capability of the transformer architecture to capture long-term temporal dependencies in traffic flow data and using MI between geographically-distributed sensors to incorporate geographical context. By computing MI between target sensors and their surrounding peers, the most informative sensors (covariates) that encode significant spatio-temporal traffic flow information are identified. Augmenting the \gattf~model’s input with these covariates enriches the feature space and improves the accuracy of the model's forecasting, without increasing the model complexity. The effectiveness of \gattf~is validated using real-world motorway traffic data from the Geneva urban motorway.

Furthermore, MI systematically reveals the intrinsic complexity of traffic-flow dynamics shaped by network topology, as shown by the analyzed sensors whose MI values logically capture the traffic dynamic dependencies between upstream and downstream motorway locations. In this way, MI provides interpretability by quantifying information sharing among sensors and enables conditioning the Transformer learning process.

So far, we have compared \gattf~against a limited set of baselines, primarily serving as an internal ablation around our architecture, to verify the effectiveness of the concept using simple and easily comparable models. However, it is possible that specific combinations of informative and less informative covariates could yield additional insights into the underlying traffic flow dynamics. Therefore, a sensitivity analysis in this direction is warranted, and well-established deep-learning architectures for spatio-temporal time series forecasting, including spatio-temporal graph neural networks, will also be examined, with the inclusion of additional context-aware exogenous variables. Future work will focus on integrating \gattf~with simulation-based digital twins to enable forecast-driven microscopic simulations that extend beyond short-horizon, real-time traffic mirroring.

\section*{Acknowledgment}

This work was supported in part by Taighde \'{E}ireann -- Research Ireland under grant number 21/FFP-A/8957, Croatian Science Foundation under grant numbers HRZZ-MOBODL-2023-12-5261 and UIP-2025-02-1083, and by the European Union -- NextGenerationEU as part of the institutional research project of the University of Zagreb, Faculty of Transport and Traffic Sciences.
For the purpose of Open Access, the author has applied a CC BY public copyright license to any Author Accepted Manuscript version arising from this submission.

\bibliographystyle{IEEEtran}
\bibliography{References}

@ARTICLE{TraffPrediction_MI_Clustering_2023_Huang,
author={Huang, Yunjie and Song, Xiaozhuang and Zhu, Yuanshao and Zhang, Shiyao and Yu, James J. Q.},
journal={IEEE Transactions on Intelligent Transportation Systems}, 
title={Traffic Prediction With Transfer Learning: A Mutual Information-Based Approach}, 
year={2023},
volume={24},
number={8},
pages={8236-8252},
doi={10.1109/TITS.2023.3266398}
}

@misc{yamaguchi2025citras_covariate_informed_transformer_time_series,
title={CITRAS: Covariate-Informed Transformer for Time Series Forecasting}, 
author={Yosuke Yamaguchi and Issei Suemitsu and Wenpeng Wei},
year={2025},
eprint={2503.24007},
archivePrefix={arXiv},
primaryClass={cs.LG},
url={https://arxiv.org/abs/2503.24007}, 
}

@misc{lim2020temporal_fusion_transformers_known_unknown_covariates,
title={Temporal Fusion Transformers for Interpretable Multi-horizon Time Series Forecasting}, 
author={Bryan Lim and Sercan O. Arik and Nicolas Loeff and Tomas Pfister},
year={2020},
eprint={1912.09363},
archivePrefix={arXiv},
primaryClass={stat.ML},
url={https://arxiv.org/abs/1912.09363}, 
}

@ARTICLE{2024_Irfan_DT_in_Transport_Review,
author={Irfan, Muhammad Sami and Dasgupta, Sagar and Rahman, Mizanur},
journal={IEEE Internet of Things Journal}, 
title={Toward Transportation Digital Twin Systems for Traffic Safety and Mobility: A Review}, 
year={2024},
volume={11},
number={14},
pages={24581-24603},
doi={10.1109/JIOT.2024.3395186}
}

@ARTICLE{2025Guo_LLM_driven_DT_for_ITS_Survey,
author={Guo, Zhiqi and Tang, Fengxiao and Luo, Linfeng and Zhao, Ming and Kato, Nei},
journal={IEEE Communications Surveys \& Tutorials}, 
title={A Survey on Applications of Large Language Model-Driven Digital Twins for Intelligent Network Optimization}, 
year={2025},
volume={},
number={},
pages={1-1},
doi={10.1109/COMST.2025.3568637}
}

@ARTICLE{Zhang2024_Traffic_Forecasting_for_Freeway_DT,
author={Zhang, Weibin and Zha, Huazhu and Gan, Lu and Li, Qianmu},
journal={IEEE Transactions on Intelligent Transportation Systems}, 
title={DS-TFSN-Based Vehicle Travel Time Prediction Method for Digital Twin System of Freeways}, 
year={2024},
volume={25},
number={12},
pages={20073-20084},doi={10.1109/TITS.2024.3451714}}

@ARTICLE{2020Cai_Traffic_Transformer,
author = {{Cai}, Ling and {Janowicz}, Krzysztof and {Mai}, Gengchen and {Yan}, Bo and {Zhu}, Rui},
title = "{Traffic transformer: Capturing the continuity and periodicity of time series for traffic forecasting}",
journal = {Transactions in GIS},
 year = 2020,
month = jun,
volume = {24},
number = {3},
pages = {736-755},
  doi = {10.1111/tgis.12644},
adsurl = {https://ui.adsabs.harvard.edu/abs/2020TrGIS..24..736C}
}

@ARTICLE{2025Chang_Trafficformer_short_term_forecasting_considering_traffic_spatiotemporal_correlation,
AUTHOR={Chang, Ande  and Ji, Yuting  and Bie, Yiming },
TITLE={Transformer-based short-term traffic forecasting model considering traffic spatiotemporal correlation},
JOURNAL={Frontiers in Neurorobotics},
VOLUME={Volume 19 - 2025},
YEAR={2025},
URL={https://www.frontiersin.org/journals/neurorobotics/articles/10.3389/fnbot.2025.1527908},
DOI={10.3389/fnbot.2025.1527908},
ISSN={1662-5218}
}

@article{2023_KUSIC_DT_GM,
title = {A digital twin in transportation: Real-time synergy of traffic data streams and simulation for virtualizing motorway dynamics},
journal = {Advanced Engineering Informatics},
volume = {55},
pages = {101858},
year = {2023},
issn = {1474-0346},
doi = {https://doi.org/10.1016/j.aei.2022.101858},
url = {https://www.sciencedirect.com/science/article/pii/S1474034622003160},
author = {Krešimir Kušić and René Schumann and Edouard Ivanjko}
}

@article{Kraskov2004_estim_mutual_info,
  title = {Estimating mutual information},
  author = {Kraskov, Alexander and St\"ogbauer, Harald and Grassberger, Peter},
  journal = {Phys. Rev. E},
  volume = {69},
  issue = {6},
  pages = {066138},
  numpages = {16},
  year = {2004},
  publisher = {American Physical Society},
  doi = {10.1103/PhysRevE.69.066138},
}

@article{Li_2023_Mitigating_urban_mot_congest_via_active_TM,
title = {Mitigating urban motorway congestion and emissions via active traffic management},
journal = {Research in Transportation Business \& Management},
volume = {48},
pages = {100789},
year = {2023},
doi = {https://doi.org/10.1016/j.rtbm.2022.100789},
author = {Duo Li and Joan Lasenby}
}

@article{SHAO2025_CCDSReFormer,
title = {CCDSReFormer: Traffic flow prediction with a criss-crossed dual-stream enhanced rectified transformer model},
journal = {Communications in Transportation Research},
volume = {5},
pages = {100189},
year = {2025},
doi = {https://doi.org/10.1016/j.commtr.2025.100189},
author = {Zhiqi Shao and Michael G.H. Bell and Ze Wang and D. Glenn Geers and Xusheng Yao and Junbin Gao}
}

@InProceedings{Zhang_2023_Pure_Transformer_for_Traff_Forecasting,
author="Zhang, Junhao
and Jin, Juncheng
and Tang, Junjie
and Qu, Zehui",
editor="Iliadis, Lazaros
and Papaleonidas, Antonios
and Angelov, Plamen
and Jayne, Chrisina",
title="FPTN: Fast Pure Transformer Network for Traffic Flow Forecasting",
booktitle="Artificial Neural Networks and Machine Learning -- ICANN 2023",
year="2023",
publisher="Springer Nature Switzerland",
address="Cham",
pages="382--393",
}

@inproceedings{2017_Vaswani_AttentionIsAllYouNeed,
 author = {Vaswani, Ashish and Shazeer, Noam and Parmar, Niki and Uszkoreit, Jakob and Jones, Llion and Gomez, Aidan N and Kaiser, \L ukasz and Polosukhin, Illia},
 booktitle = {Advances in Neural Information Processing Systems},
 editor = {I. Guyon and U. Von Luxburg and S. Bengio and H. Wallach and R. Fergus and S. Vishwanathan and R. Garnett},
 pages = {},
 publisher = {Curran Associates, Inc.},
 title = {Attention is All you Need},
 url = {https://proceedings.neurips.cc/paper_files/paper/2017/file/3f5ee243547dee91fbd053c1c4a845aa-Paper.pdf},
 volume = {30},
 year = {2017}
}

@article{freedman1981histogram,
  title={On the histogram as a density estimator: L2 theory},
  author={Freedman, David and Diaconis, Persi},
  journal={Zeitschrift f{\"u}r Wahrscheinlichkeitstheorie und Verwandte Gebiete},
  volume={57},
  number={4},
  pages={453--476},
  year={1981},
  publisher={Springer}
}

@misc{2020_wolf_huggingfacestransformersstateoftheartnatural,
title={HuggingFace's Transformers: State-of-the-art Natural Language Processing}, 
author={Thomas Wolf and Lysandre Debut and Victor Sanh and Julien Chaumond and Clement Delangue and Anthony Moi and Pierric Cistac and Tim Rault and Rémi Louf and Morgan Funtowicz and Joe Davison and Sam Shleifer and Patrick von Platen and Clara Ma and Yacine Jernite and Julien Plu and Canwen Xu and Teven Le Scao and Sylvain Gugger and Mariama Drame and Quentin Lhoest and Alexander M. Rush},
year={2020},
eprint={1910.03771},
archivePrefix={arXiv},
primaryClass={cs.CL},
url={https://arxiv.org/abs/1910.03771}, 
}

@inproceedings{Qingsong_2023_Transformers_in_time_series_survey,
author = {Wen, Qingsong and Zhou, Tian and Zhang, Chaoli and Chen, Weiqi and Ma, Ziqing and Yan, Junchi and Sun, Liang},
title = {Transformers in time series: a survey},
year = {2023},
isbn = {978-1-956792-03-4},
url = {https://doi.org/10.24963/ijcai.2023/759},
doi = {10.24963/ijcai.2023/759},
booktitle = {Proceedings of the Thirty-Second International Joint Conference on Artificial Intelligence},
articleno = {759},
numpages = {9},
series = {IJCAI '23}
}

@article{SHAYGAN2022_TrafficPrediction_using_AI_review,
title = {Traffic prediction using artificial intelligence: Review of recent advances and emerging opportunities},
journal = {Transportation Research Part C: Emerging Technologies},
volume = {145},
pages = {103921},
year = {2022},
issn = {0968-090X},
doi = {https://doi.org/10.1016/j.trc.2022.103921},
author = {Maryam Shaygan and Collin Meese and Wanxin Li and Xiaoliang (George) Zhao and Mark Nejad},
}

@ARTICLE{2023Fan_CAV_DT,
author={Fan, Bo and Su, Zixun and Chen, Yanyan and Wu, Yuan and Xu, Chengzhong and Quek, Tony Q. S.},
journal={IEEE Wireless Communications}, 
title={Ubiquitous Control Over Heterogeneous Vehicles: A Digital Twin Empowered Edge AI Approach}, 
year={2023},
volume={30},
number={1},
pages={166-173},
doi={10.1109/MWC.012.2100587}
}

\end{document}